\documentclass[11pt]{article}
\pdfoutput=1
\PassOptionsToPackage{table,dvipsnames}{xcolor}
\usepackage[final]{acl}

\usepackage{times}
\usepackage{latexsym}

\usepackage[T1]{fontenc}

\usepackage[utf8]{inputenc}

\usepackage{microtype}

\usepackage{inconsolata}

\usepackage{multirow}
\usepackage{graphicx}
\usepackage{subfigure}
\usepackage{booktabs}
\usepackage{nicematrix}
\usepackage[capitalise,noabbrev]{cleveref}
\usepackage{adjustbox}
\usepackage{hyperref}
\usepackage{cmatrix}
\usepackage{enumitem}
\usepackage{comment}
\providecommand\new{{\small \textcolor{red}{\textbf{[new]}}}}

\title{Latxa: An Open Language Model and Evaluation Suite for Basque}

\author{Julen Etxaniz\thanks{Equal contribution.} \quad Oscar Sainz\footnotemark[1] \quad Naiara Perez\footnotemark[1] \quad Itziar Aldabe \quad German Rigau \\
\bf Eneko Agirre \quad Aitor Ormazabal \quad Mikel Artetxe \quad Aitor Soroa \\
HiTZ Center - Ixa, University of the Basque Country UPV/EHU \\
\texttt{\{julen.etxaniz,a.soroa\}@ehu.eus} \\
}

\begin{document}
\maketitle
\begin{abstract}
We introduce Latxa, a family of large language models for Basque ranging from 7 to 70 billion parameters. 
Latxa is based on Llama 2, which we continue pretraining on a new Basque corpus comprising 4.3M documents and 4.2B tokens. Addressing the scarcity of high-quality benchmarks for Basque, we further introduce 4 multiple choice evaluation datasets: EusProficiency, comprising 5,169 questions from official language proficiency exams; EusReading, comprising 352 reading comprehension questions; EusTrivia, comprising 1,715 trivia questions from 5 knowledge areas; and EusExams, comprising 16,774 questions from public examinations. In our extensive evaluation, Latxa outperforms all previous open models we compare to by a large margin. In addition, it is competitive with GPT-4 Turbo in language proficiency and understanding, despite lagging behind in reading comprehension and knowledge-intensive tasks. Both the Latxa family of models, as well as our new pretraining corpora and evaluation datasets, are publicly available under open licenses.\footnote{\url{https://github.com/hitz-zentroa/latxa}} Our suite enables reproducible research on methods to build LLMs for low-resource languages.

\end{abstract}


\section{Introduction}

\begin{table*}[t]
\centering
\small
\begin{adjustbox}{max width=\linewidth}
\begin{NiceTabular}{rrc*{7}{c}}
\CodeBefore
\columncolor{gray!5}{10}
\rowcolor{purple!5}{10}
\rowcolor{purple!5}{13}
\rowcolor{purple!5}{17}
\Body
\toprule
&&&&&\new&\new&\new&\new& \\
  && \textbf{XStory} & \textbf{Belebele} & \textbf{BasGLUE} & \textbf{EusProf} & \textbf{EusRead} & \textbf{EusTrivia} & \textbf{EusExams} & \textbf{Avg} \\
  \midrule
  \textbf{Random} & & 50.00 & 25.00 & 37.50 & 25.00 & 25.83 & 26.55 & 25.00 & 30.70 \\
 \midrule
 \textbf{GPT-3.5 Turbo} & n/a
                &   --    & 57.33 & 48.62 & 31.24 & 36.65 & 46.71 &  42.42 & --    \\
 \textbf{GPT-4 Turbo} & n/a
                &   -- & \textbf{\underline{90.67}} & \textbf{\underline{66.54}} & \textbf{56.70} &   \textbf{\underline{75.85}}    & \textbf{\underline{73.12}} & \textbf{\underline{70.22}} & --       \\
\midrule
 \textbf{XGLM} & 7B      & 57.71 & 23.88 & 41.47 & 22.96 & 24.43 & 26.53 & 24.59 & 31.65 \\
 \textbf{BLOOM} & 7B     & 57.18 & 27.00 & 40.17 & 25.34 & 28.41 & 27.17 & 25.07 & 32.91 \\
 \textbf{Mistral} & 7B   & 51.09 & \textbf{38.89} & 39.22 & 25.01 & \textbf{29.26} & 34.58 & 32.15 & 35.74 \\
\textbf{Llama 2} & 7B           & 50.43 & 26.22 & 38.20 & 24.09 & 27.27 & 29.50 & 28.84 & 32.08 \\
\new{} \textbf{Latxa} & 7B           & \textbf{65.45} & 37.33 & \textbf{52.56} & \textbf{30.26} & 25.00 & \textbf{42.16} & \textbf{33.82} & \textbf{40.94} \\

\midrule
 \textbf{mGPT} & 13B     & 55.39 & 25.00 & 37.56 & 25.00 & 24.15 & 27.17 & 25.73 & 31.43 \\
\textbf{Llama 2} & 13B          & 50.63 & 32.00 & 38.98 & 25.90 & 28.98 & 33.53 & 29.66 & 34.24 \\
\new{} \textbf{Latxa} & 13B          & \textbf{66.51} & \textbf{53.89} & \textbf{53.36} & \textbf{44.11} & \textbf{32.67} & \textbf{56.38} & \textbf{43.66} & \textbf{50.08} \\
\midrule
 \textbf{Mixtral} & 8x7B & 52.55 & 50.44 & 45.00 & 26.43 & 37.50 & 42.51 & 39.87 & 42.04 \\
 \textbf{Yi}      & 34B  & 52.22 & 54.56 & 43.90 & 27.30 & 34.66 & 42.57 & 39.68 & 42.13 \\
\textbf{Llama 2} & 70B          & 51.62 & 33.56 & 42.55 & 24.16 & 27.84 & 38.43 & 33.08 & 35.89 \\
\new{} \textbf{Latxa} & 70B          & \textbf{\underline{70.55}} & \textbf{71.67} & \textbf{59.74} & \textbf{\underline{60.65}} & \textbf{50.57} & \textbf{62.45} & \textbf{51.90} & \textbf{\underline{61.08}}      \\
\bottomrule
\end{NiceTabular}
\end{adjustbox}
\caption{\textbf{Main results.} Best results in each compute class are in \textbf{bold}. Best overall results are \underline{underlined}.}
\label{tab:intro-summary}
\end{table*}

Motivated by their increasing training cost and commercial interest, the development of Large Language Models (LLMs) has been led by close initiatives like GPT~\cite{openai2023gpt4}, Claude~\cite{wu2023comparative} and Gemini~\cite{geminiteam2023gemini}. In recent times, a more open ecosystem has emerged following the release of various competitive models like Llama 2 \cite{touvron2023Llama} and Mistral~\cite{jiang2024mixtral}. However, despite early efforts to build open multilingual models~\cite{lin-etal-2022-shot,workshop2023bloom}, the most competitive ones are notoriously English-centric. As shown in Table~\ref{tab:intro-summary}, all these open models perform poorly in low-resource languages like Basque, with most results marginally surpassing random chance.

In this work, we present Latxa, an open family of LLMs for Basque that substantially outperforms all these previous models. 
Basque is an agglutinative language written in Latin script and with no known relatives, although a significant part of the vocabulary is shared with contact languages like Spanish and French. Basque is the 52th language in Common Crawl, with 0.035\% of the total content --for reference, English is the 1st language with 46\% of the content and Spanish is the 5th with 4.6\%.
Our work builds on various open resources and models that we further expand to Basque, highlighting the importance of an open ecosystem for the development of language technology for low-resource languages. In particular, our models are based on Llama 2, which we continue training in Basque using a new corpus with 4.3M documents from 4 existing and 3 new sources. In addition, we release 4 diverse and challenging multiple-choice benchmarks comprising a total of 23,282 questions, covering language proficiency, reading comprehension, trivia questions, and public examinations.


As shown in Table \ref{tab:intro-summary}, Latxa performs substantially better than all existing open models, with the 70B variant outperforming the previous best open model (Yi 34B) by 18.95 points in average. In addition, it also outperforms the Llama 2 model it is based on by 25.18 points, and it is also superior to GPT-3.5 Turbo in all datasets we evaluate on. Interestingly, our best model also outperforms GPT-4 Turbo in language proficiency exams (EusProf), despite lagging behind in reading comprehension and knowledge-intensive tasks. This suggests that the capabilities that an LLM exhibits in a given language are not determined by its linguistic competence in this particular language, opening the doors to further improvements in low-resource LLMs as stronger English models become available.

This paper makes the following contributions:  \textbf{(1)} We release a high-quality corpus for Basque, comprising 4.3M documents and 4.2B tokens. The corpus combines the EusCrawl v1.1, Egunkaria, Booktegi, Wikipedia, CulturaX, Colossal OSCAR and HPLT v1 datasets (the first 3 being new), which we carefully deduplicate and filter. \textbf{(2)} We release the Latxa family of Basque LLMs, comprising 3 models with 7B, 13B and 70B parameters. \textbf{(3)} We release 4 new multiple-choice benchmarks for Basque: EusProficiency (official language proficiency exams), EusReading (reading comprehension), EusTrivia (trivia questions from 5 knowledge areas), and EusExams (public examinations). \textbf{(4)} We present extensive experiments comparing Latxa to previous open and closed models. \textbf{(5)} We show that it is possible to train significantly stronger LLMs for low-resource languages building on the existing ecosystem of open models and resources. In a similar spirit to other open LLMs, such as Pythia \citep{biderman2023pythia}, LLM360 \citep{liu2023llm360} and OLMO \citep{groeneveld2024olmo}, we release all the necessary data, code, weights and documentation to run and evaluate our models, facilitating similar efforts for other low-resource languages.

\section{Training Corpora} \label{sec:corpora}

Our training corpus combines various existing datasets, as well as some new ones that we release with this work. We have prioritized quality over quantity when constructing our corpus, prioritizing high-quality data sources and applying a thorough deduplication and filtering process. 
We next describe our data sources in \S\ref{ssec:data-sources}, followed by our preprocessing pipeline in \S\ref{ssec:data-preprocessing}. Table \ref{tab:data} summarizes the statistics of the resulting dataset.





\subsection{Data Sources}
\label{ssec:data-sources}



\begin{table*}[t]
    \small
    \centering
    \begin{NiceTabular}{r*{7}{r}l}
        \CodeBefore
        \columncolor{gray!5}{6,7,8}
        \rowcolor{purple!5}{4,8,9}
        \Body
        \toprule
                       & \multicolumn{2}{c}{\textbf{Raw}} & \multicolumn{2}{c}{\textbf{Deduped}} & \multicolumn{3}{c}{\textbf{Filtered}} \\
        \cmidrule(lr){2-3} \cmidrule(lr){4-5} \cmidrule(lr){6-8}
                       & \textbf{Docs} & \textbf{Words}   & \textbf{Docs} & \textbf{Words} & \textbf{Docs} &  \textbf{Words} & \textbf{Toks} & \textbf{Source} \\
        \midrule
        \textbf{CulturaX}       & 1.60M &  622M & 1.33M &  548M & 1.31M &  541M & 1.84B & {\href{https://hf.co/datasets/uonlp/CulturaX}{\texttt{hf.co/uonlp/CulturaX}}} \\
        \new{} \textbf{EusCrawl v1.1}  & 2.12M &  411M & 1.94M &  384M & 1.79M &  359M & 1.21B & {\citet{artetxe-etal-2022-corpus}}\\
        \textbf{HPLT v1}        & 2.29M & 1.55B & 1.56M &  312M & 0.37M &  120M & 421M & {\href{https://hplt-project.org}{\texttt{hplt-project.org}}}\\
        \textbf{Colossal OSCAR}       & 0.65M &  283M & 0.25M &  111M & 0.24M &  105M & 380M & {\href{https://hf.co/datasets/oscar-corpus/colossal-oscar-1.0}{\texttt{hf.co/oscar-corpus}}} \\
        \textbf{Wikipedia}         & 0.55M &   54M & 0.55M &   54M & 0.41M &   51M & 182M & {\href{https://dumps.wikimedia.org}{\texttt{dumps.wikimedia.org}}} \\
        \new{} \textbf{Egunkaria}      & 0.18M &   40M & 0.18M &   39M & 0.18M &   39M & 129M & {n/a}\\
        \new{} \textbf{Booktegi}       &   181 &    3M &   177 &    3M &   166 &    3M & 8M & {\href{https://www.booktegi.eus}{\texttt{booktegi.eus}}} \\
        \midrule
        \textbf{Total}          & 7.39M & 2.96B & 5.80M & 1.45B & 4.30M & 1.22B & 4.17B & \\
        \bottomrule
    \end{NiceTabular}
    \caption{Data sources and statistics at each preprocessing stage. \textit{``Toks''} are Llama 2 tokens.}
    \label{tab:data}
\end{table*}

\noindent \textbf{\new{} EusCrawl v1.1.} The original version of EusCrawl \cite{artetxe-etal-2022-corpus} was built using ad-hoc scrapers to extract text from 33 newswire websites, resulting in higher quality documents compared to general-purpose approaches. In this work, we release an updated version of EusCrawl, including new content up to November 2023. This increases the number of unique documents from 1.38 to 1.94 millions, for a total of 384 million words.

\noindent \textbf{\new{} Egunkaria.} Euskaldunon Egunkaria was a daily newspaper written fully in the Basque language. The corpus includes approximately 176k news articles, editorials, and various types of reviews from the years 2001 to 2006, totalling 39 million words.

\noindent \textbf{\new{} Booktegi.} The Booktegi platform hosts free content in Basque, such as books, interviews, and audio materials. The corpus comprises approximately 3 million words from 166 EPUB books covering essays, fiction, and poetry.

\noindent \textbf{Wikipedia.} We download and process a Basque Wikipedia dump,\footnote{The \texttt{20231101} dump corresponding to November 2023.} obtaining nearly 550k documents and more than 54 million words. The plain text has been extracted with WikiExtractor.\footnote{\url{https://github.com/attardi/wikiextractor}}

\noindent \textbf{CulturaX.} CulturaX \citep{nguyen2023culturax} is a large multilingual dataset resulting from the combination and processing of mC4 \citep{xue-etal-2021-mt5} and the four OSCAR releases \texttt{2019}, \texttt{21.09}, \texttt{22.01}, and \texttt{23.01} \citep{OrtizSuarezSagotRomary2019}. These corpora originate, in turn, from 66 Common Crawl (CC) snapshots spanning from 2015 to 2022. Basque content constitutes 0.02\% of CulturaX, encompassing nearly 1.60 million documents and 622 million words.

\noindent \textbf{Colossal OSCAR.} The largest release of the OSCAR project \citep{OrtizSuarezSagotRomary2019} to date, Colossal OSCAR 1.0, is based on 10 CC snapshots. Here, we use the two snapshots not covered by CulturaX, namely, \texttt{06-07-22} and \texttt{05-06-23}. Additionally, we have had access to an OSCAR-processed CC snapshot from April 2023. In total, we have obtained almost 650k documents in Basque from these datasets, totalling 282 million words.

\noindent \textbf{HPLT v1.} The High Performance Language Technologies project \citep[HPLT;][]{aulamo-etal-2023-hplt} compiled another massive, multilingual dataset from the Internet Archive and CC. In this work, we use the first release, which contains 2.29 million documents (1.55 billion words) in Basque. It must be noted that, unlike the aforementioned web-based sources, the HPLT dataset was released without any deduplication or filtering. Consequently, our preprocessing approach has been particularly aggressive with this dataset (see \S\ref{ssec:data-preprocessing}).


\subsection{Preprocessing}
\label{ssec:data-preprocessing}

We used the Dolma toolkit \cite{soldaini2024dolma} and Corpus Cleaner v2 \citep[CCv2;][]{palomar-giner-etal-2024-curated-catalog} to normalize, deduplicate and filter the datasets. Since the majority of our data sources were intentionally selected, organized, and/or curated by their respective authors, our main focus has been on removing outliers and cross-dataset duplicates. This process is briefly outlined below, with further details available in \cref{app:data}.  The final size of the processed corpus is shown in Table \ref{tab:data}. In total, it amounts to 1.22B words and 4.17B Llama 2 tokens. Each dataset is shuffled and then split separately into testing (1\%), development (1\%) and training (98\%) example sets. 

\noindent \textbf{Normalization.} CCv2 is first used to fix document encoding and whitespace normalization.

\noindent \textbf{Deduplication.} Cross-dataset document repetitions are identified and removed at both the URL and document content levels. Specifically, we conduct near-deduplication with Bloom filters \cite{bloom1970space} as implemented in Dolma. To maximise corpus quality, we prioritized content from well-curated sources (Wikipedia, EusCrawl, Egunkaria and Booktegi) then from massive but comparatively cleaner sources (CulturaX and Colossal OSCAR) over HPLT (see Figure \ref{fig:data-a} in Appendix \ref{app:data}). The latter undergoes additional deduplication at the paragraph level.

\noindent \textbf{Filtering.} Documents unlikely to contain quality content are identified and removed in two stages. First, a combination of heuristics from Gopher \cite{rae2022scaling} and C4 \cite{raffel2020exploring}, with adaptations tailored to the Basque language is applied (e.g.,  regarding average word length). We also perform language identification with CLD2 through Dolma, which has predominantly impacted HPLT, with approximately one-third of this corpus being discarded at this stage. Finally, the corpora are processed with CCv2, which assigns an aggregated quality score per document based on a comprehensive set of taggers. Once again, HPLT has been affected most, resulting in a further 25\% reduction in document counts for this dataset.

\section{Latxa Models}
\label{sec:models}

We train 7B, 13B and 70B models following a continued pretraining approach. To that end, we use Llama 2 as the base model~\cite{touvron2023Llama}, and continue training it using the corpus described in \S\ref{sec:corpora}. To mitigate catastrophic forgetting from the original model\footnote{Our preliminary experiments showed that adding English data was critical for strong few-shot performance. For instance, an early version of our 7B model without English data obtained 23.67 points on BeleBele \citep{bandarkar2023belebele}, 13 points less than the corresponding version with English data.}, we also include English data in the continued pretraining stage. For that purpose, we use 500k random documents from The Pile~\cite{gao2020pile}, totaling 0.9B tokens. 

\subsection{Pretraining Details}

\begin{figure}
    \centering
    \includegraphics[width=\linewidth]{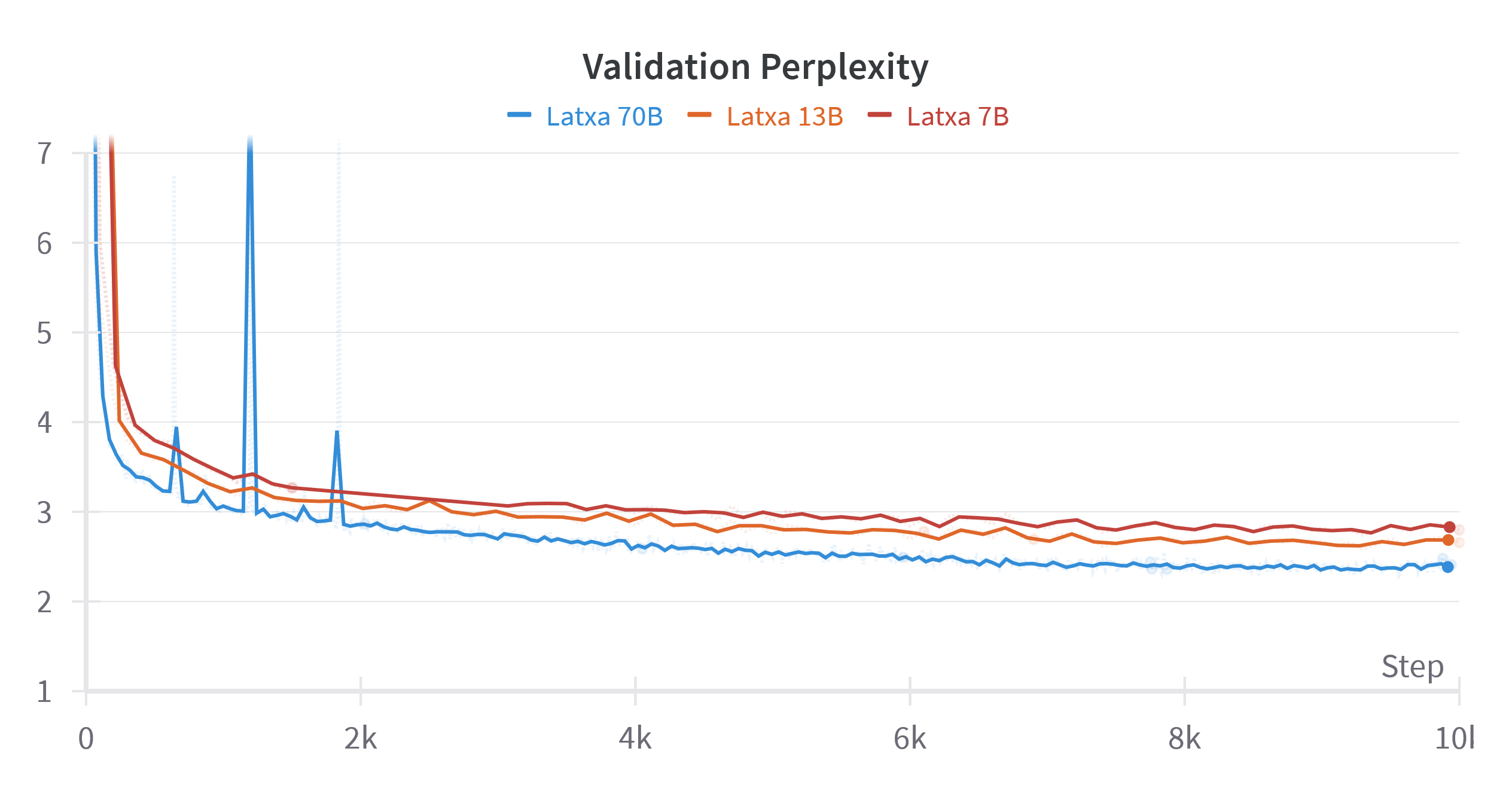}
    \caption{Validation perplexity throughout training.}
    \label{fig:validation_perp}
\end{figure}

The training of Latxa has been conducted using the GPT-Neox~\cite{gpt-neox-library} library. As infrastructure, we have leveraged the CINECA HPC Leonardo computing cluster located in Italy, which is powered by 3,456 nodes each containing 4x custom A100 64GB GPUs. 
The models are trained for 10k steps with a sequence length of 4,096 tokens and an effective batch size of 1M tokens, resulting in a total of 10B tokens. 
We use a cosine learning rate schedule, with a warm-up of 500 steps and decaying down to 3\% of the peak learning rate. We set up the peak learning rate to be $1\times10^{-4}$. All other hyperparameters follow ~\citet{touvron2023Llama}.
Figure~\ref{fig:validation_perp} shows the validation perplexity during training. 












\subsection{Carbon Emissions}


\begin{table}
    \centering
    \small
    \begin{adjustbox}{max width=\linewidth}
    \begin{tabular}{rrr}
        \toprule
         & \multicolumn{1}{c}{\textbf{Time}} & \textbf{Carbon Emitted} \\
         \textbf{Size} & \textbf{(GPU Hours)} & \multicolumn{1}{c}{(\textbf{kg CO\textsubscript{2} eq})} \\ \midrule
          7B & 952.53h & 124.47kg \\ 
         13B & 2,518.0h & 329.06kg \\ 
          70B & 30,266.0h & 3,955.17kg \\ \midrule 
         Total & 33,636.5h & 4,408.7kg \\ \bottomrule
    \end{tabular}
    \end{adjustbox}
    \caption{Carbon footprint of training different models}
    \label{tab:carbon_emissions}

\end{table}

Pretraining LLMs requires compute-expensive experiments, carrying a significant carbon footprint. The 7B, 13B and 70B models were trained on 32, 64 and 256 GPUs, respectively. We report the compute hours and power consumption involved in our experiments in Table~\ref{tab:carbon_emissions}. The carbon emitted was estimated using a GPU power consumption of $440W$ and a carbon efficiency of $0.297 kg/kWh$ (carbon efficiency on Italy on February 9, 2024, according to ElectricityMaps\footnote{\url{https://www.electricitymaps.com/}}).


\section{New Evaluation Datasets}
\label{sec:new-datasets}

To overcome the scarcity of Basque benchmarks that are suitable for evaluating base language models, we collect new evaluation data from various online games and tests. We have decided to take this approach instead of translating existing datasets to avoid translation artifacts \citep{artetxe-etal-2020-translation}. Most importantly, this allows to have localized datasets that test the models' knowledge about topics that are most relevant for Basque speakers. These tasks cover language proficiency (EusProficiency), reading comprehension (EusReading), trivia questions (EusTrivia), and exams of advanced professional level (EusExams). All the datasets consist of multiple-choice questions, making them suitable for few-shot learning akin to MMLU \citep{hendrycks2021measuring} in English. We next describe each dataset in more detail, while \cref{tab:eval-data} summarizes their statistics. For examples of each task, see \cref{tab:eval-framework-eus} in \cref{app:examples_prompts}.

\paragraph{EusProficiency.}
EusProficiency comprises 5,169 exercises on different topics from past EGA exams, the official C1-level certificate of proficiency in Basque.
We have collected the \textit{atarikoa} exercises from EGA exams through the years 1998 to 2008. Atarikoa is the first qualifying test of EGA, which measures different aspects of language competency, such as reading comprehension, grammar, vocabulary, spelling, and writing. Each test generally has 85 multiple-choice questions, with 4 choices and a single correct answer. Currently, there is no comparable dataset available, nor could one be obtained by translating existing analogous datasets from other languages.

\paragraph{EusReading.}
EusReading consists of 352 reading comprehension exercises (\textit{irakurmena}) sourced from the same set of past EGA exams.
Each test generally has 10 multiple-choice questions, with 4 choices and a single correct answer. These exercises are more challenging than Belebele \citep{bandarkar2023belebele} due to the complexity and length of the input texts (see \cref{tab:eval-data}). As a result, EusReading is useful to measure long context understanding of models.

\paragraph{EusTrivia.}
EusTrivia consists of 1,715 trivia questions from multiple online sources. 56.3\% of the questions are elementary level (grades 3-6), while the rest are considered challenging. A significant portion of the questions focuses specifically on the Basque Country, its language, and culture. Each multiple-choice question contains two, three or four choices (3.84 on average) and a single correct answer. Five areas of knowledge are covered:

\begin{itemize}[noitemsep]
    \item \textbf{Humanities and Natural Sciences} (27.8\%): This category encompasses questions about history, geography, biology, ecology and other social and natural sciences.
    \item \textbf{Leisure and Art} (24.5\%): This category includes questions on sports and athletes, performative and plastic arts and artists, architecture, cultural events, and related topics.
    \item \textbf{Music} (16.0\%): Here are grouped all the questions about music and musicians, both classical and contemporary.
    \item \textbf{Language and Literature} (17.1\%): This category is concerned with all kinds of literature productions and writers, as well as metalinguistic questions (e.g., definitions, synonyms, and word usage).
    \item \textbf{Mathematics and ICT} (14.5\%): This category covers mathematical problems and questions about ICT, as well as questions about people known for their contributions to these fields of knowledge.
\end{itemize}

\paragraph{EusExams.}
EusExams is a collection of tests designed to prepare individuals for Public Service examinations conducted by several Basque institutions, including the public health system Osakidetza, the Basque Government, the City Councils of Bilbao and Gasteiz, and the University of the Basque Country (UPV/EHU). Within each of these groups, there are different exams for public positions, such as administrative and assistant roles.
Each multiple-choice question contains 2 to 4 choices (3.90 on average) and one correct answer. The dataset is mostly parallel with 16k questions in Basque and 18k in Spanish, from which we only consider the Basque subset.
It could be said to be similar to MMLU’s professional-level questions, but focusing on knowledge relevant to the Basque community, with questions related to local public services and law.

\begin{table}[t]
    \centering
    \small
    \addtolength{\tabcolsep}{-2pt}
    \begin{NiceTabular}{rrrrr}
    \CodeBefore
    \rowcolor{purple!5}{3,4,5,6}
    \Body
    \toprule
                     &                & \textbf{Input} & \multicolumn{2}{c}{\textbf{Output}} \\
                     \cmidrule(lr){3-3} \cmidrule(lr){4-5}
                     & \textbf{Items} & \textbf{chars} & \textbf{choices}           & \textbf{chars} \\
    \midrule
    \new{} \textbf{EusProf}    & 5,169          & 50          & 4                   & 28         \\
    \new{} \textbf{EusReading} & 352            & 5,340       & 2-4                 & 67         \\
    \new{} \textbf{EusTrivia}  & 1,715          & 55          & 2-4                 & 14         \\
    \new{} \textbf{EusExams}   & 16,774         & 115         & 4                   & 63         \\
    \midrule
    \textbf{XStoryCloze}      & 1,511          & 202         & 2                    & 44         \\
    \textbf{Belebele}         & 900            & 584         & 4                    & 28         \\
    \midrule
    \textbf{BEC}              & 1,302          & 97          & 3                    &            \\
    \textbf{BHTCv1}           & 1,854          & 265         & 12                   &            \\
    \textbf{Korref}           & 587            & 275         & 2                    &            \\
    \textbf{QNLI$_{eu}$}      & 238            & 158         & 2                    &            \\
    \textbf{VaxxStance}       & 312            & 209         & 3                    &            \\
    \textbf{WiC$_{eu}$}       & 1,400          & 375         & 2                    &            \\
    \bottomrule
    \end{NiceTabular}
    \caption{Evaluation dataset statistics: number of examples, average input and output length in characters, and number of choices per example. BasqueGLUE tasks (lower section) do not have output length because they are classification tasks.}
    \label{tab:eval-data}
\end{table}

\section{Experimental Setting} 
\label{sec:experimental-setting}

To assess the quality of Latxa models, we thoroughly evaluate them on a suite of diverse and challenging tasks against state-of-the-art models. In this section, we present the baseline models (\S\ref{ssec:baselines}) and tasks (\S\ref{ssec:evaluation-dasets}), and describe the evaluation framework (\S\ref{ssec:evaluation-framework}). Results are analyzed and discussed in the next section (\S\ref{sec:results}).

\subsection{Baseline Models}
\label{ssec:baselines}


We test the Llama 2 models \cite{touvron2023Llama} to show the difference after our continued pretraining. We also evaluated other multilingual base models that do not intentionally include Basque in pretraining, namely, Mistral-7B \citep{jiang2023mistral}, Mixtral-8x7B \citep{jiang2024mixtral}, and 01.AI's Yi-34B \citep{ai2024yi}.

Furthermore, we evaluate some of the leading open multilingual language models for Basque available to date, including BLOOM-7B \citep{workshop2023bloom}, XGLM-7B \citep{lin-etal-2022-shot}, and mGPT-13B \citep{shliazhko2022mgpt}. These models cover a broader range of languages than more recent models, but are trained on fewer tokens and exhibit generally weaker performance.

Finally, we tested the latest GPT-3.5 Turbo (\texttt{gpt-3.5-turbo-0125}) and GPT-4 Turbo (\texttt{gpt-4-1106-preview} for BasqueGLUE tasks and \texttt{gpt-4-0125-preview} for the rest), as they are the leading commercial models for Basque.


\subsection{Evaluation Datasets}
\label{ssec:evaluation-dasets}

In addition to the new evaluation datasets introduced in \S\ref{sec:new-datasets}, the models have been evaluated on the following benchmarks:

\begin{itemize}[noitemsep]
    \item \textbf{Belebele} \citep{bandarkar2023belebele}: A multiple-choice reading comprehension dataset spanning 122 language variants.
    \item \textbf{XStoryCloze} \citep{lin-etal-2022-shot}: A professionally translated version of the StoryCloze dataset \citep{mostafazadeh2017lsdsem} to 10 non-English languages. StoryCloze is a commonsense reasoning dataset that consists in choosing the correct ending to a four-sentence story.
    \item \textbf{BasqueGLUE} \citep{urbizu-etal-2022-basqueglue}: A collection of 6 NLU datasets for Basque: sentiment analysis (BEC), stance detection (VaxxStance), topic classification (BTHCv2), coreference detection (EpecKorrefBin), question-answering NLI (QNLI$_{eu}$), and word-in-context (WiC$_{eu}$).
\end{itemize}

\noindent Collectively, these datasets allow us to evaluate the performance of the models on a wide range of competences including world knowledge, linguistic knowledge, reading comprehension, and common sense reasoning. 

Following previous work \cite{brown2020language,touvron2023Llama}, we check for n-gram overlaps between these evaluation datasets and Latxa's training corpus, and find no evidence of wholesale or annotation contamination \cite{palm2023aakanksha,sainz-etal-2023-nlp}. Further information on our contamination study can be consulted in \cref{app:contamination}.

\subsection{Evaluation Framework}
\label{ssec:evaluation-framework}

The models are evaluated using the LM Evaluation Harness library \citep{biderman2024lessons} by Eleuther AI. To that end, we have implemented the new evaluation datasets following similar multiple-choice datasets that are already included in the library, such as Belebele. The specific prompts and examples for each task are reported in \cref{tab:eval-framework-eus} in \cref{app:examples_prompts}. BasqueGLUE has also been implemented as a generative evaluation dataset (see \cref{tab:eval-framework-basqueglue}).\footnote{Where possible, we based our prompts on existing analogous datasets: \href{https://github.com/EleutherAI/lm-evaluation-harness/tree/main/lm_eval/tasks/glue/qnli}{QNLI} for QNLI$_{eu}$, \href{https://github.com/EleutherAI/lm-evaluation-harness/tree/main/lm_eval/tasks/super_glue/wic}{WiC} for WiC$_{eu}$, and \href{https://github.com/EleutherAI/lm-evaluation-harness/tree/main/lm_eval/tasks/super_glue/wsc}{WSC} for EpecKorrefBin.}

We use 5 in-context examples for all tasks but two: following common practice, XStoryCloze is evaluated in a 0-shot setting, and EusReading is evaluated in a 1-shot fashion, as more examples would not fit into the context of most models.
In all cases, we compute the log probabilities of all candidates, and pick the one with the highest score as the models' final answer.

For GPT models, we have implemented the evaluation using the OpenAI API. We have kept the evaluation as similar as possible to allow a fair comparison with our models. As getting log probabilities of candidate XStoryCloze continuations from the API is not possible, we have decided not to evaluate GPT in that task. 
For few-shot tasks, we use the same prompts and provide few-shot examples as user and assistant messages. In addition, we use a system prompt in English to specify the set of candidate answers per task (see \cref{app:examples_prompts}). 

\section{Results} 
\label{sec:results}

We report our main results in \cref{tab:intro-summary}, while \cref{tab:results-eustrivia} and \cref{tab:basqueglue} report fine-grained results on the different subsets of EusTrivia and BasqueGLUE, respectively. In what follows, we summarize our main findings:

\begin{table}[t]
\addtolength{\tabcolsep}{-2.5pt}
\centering
\small
\begin{adjustbox}{max width=\linewidth}
\begin{tabular}{rr c c c c c}


\toprule

  &  & \textbf{Hum} & \textbf{Leis} & & \textbf{Lang} & \textbf{Math} \\
  &  & \textbf{\&} & \textbf{\&} & \textbf{Music} & \textbf{\&} & \textbf{\&} \\
  &  & \textbf{Nat} & \textbf{Art} & \textbf{} & \textbf{Lit} & \textbf{ICT} \\ \midrule
\textbf{GPT-3.5 Turbo} & n/a &  50.10&  48.93&  43.64&  43,00&  44.18\\
\textbf{GPT-4 Turbo} & n/a   &  \textbf{\underline{75.47}}&  \textbf{\underline{75.30}}&  \textbf{\underline{61.82}}&  \textbf{66.89}&  \textbf{\underline{84.74}}\\ 
\midrule
\textbf{XGLM} & 7B    &  23.06&  25.65&  28.73&  28.67&  29.72\\
\textbf{BLOOM} & 7B   &  22.01&  25.42&  26.91&  32.42&  \textbf{34.14}\\
\textbf{Llama 2} & 7B &  29.77&  26.84&  32.73&  32.76&  26.10\\
\rowcolor{purple!5} \new{} \textbf{Latxa} & 7B   &  \textbf{43.81}&  \textbf{41.33}&  \textbf{45.09}& \textbf{ 45.05}&  33.73 \\
\midrule
\textbf{mGPT} & 13B    &  22.22&  26.13&  26.91&  32.76&  32.13\\ 
\textbf{Llama 2} & 13B &  30.82&  31.59&  37.09&  37.88&  32.93\\
\rowcolor{purple!5} \new{} \textbf{Latxa} & 13B   &  \textbf{59.53}&  \textbf{59.85}&  \textbf{53.81}&  \textbf{62.11}&  \textbf{40.56}\\
\midrule
\textbf{Mixtral} & 8x7B &  46.54&  43.94&  40.73&  38.57&  38.96\\
\textbf{Yi}      & 34B  &  40.88&  42.99&  40.73&  41.30&  \textbf{48.59}\\ 
\textbf{Llama 2} & 70B  &  41.30&  38.24&  38.55&  35.15&  36.95\\ 
\rowcolor{purple!5} \new{} \textbf{Latxa}   & 70B  &  \textbf{67.50}&  \textbf{63.50}&  \textbf{57.81}&  \textbf{\underline{70.30}}&  46.58 \\ 
\bottomrule
\end{tabular}
\end{adjustbox}
\caption{EusTrivia results by category (accuracy).}
\label{tab:results-eustrivia}
\end{table}

\begin{table*}[t]
\centering
\small
\begin{NiceTabular}{rr*{7}{c}}
\CodeBefore
\columncolor{gray!5}{9}
\Body
\toprule
                             &  & \textbf{BEC} & \textbf{Vaxx} & \textbf{BHTC} & \textbf{Korref} & \textbf{QNLI} & \textbf{WiC} & \textbf{Avg} \\
                &               & \textit{F1}& \textit{F1}* & \textit{F1} & \textit{acc} & \textit{acc} & \textit{acc} & \\
\midrule
 \textbf{Random} && 33.33 & 33.33 & 8.33 & 50.00 & 50.00 & 50.00 & 37.50\\
 \midrule
\textbf{BERTeus}$^{\dagger}$ & 110M & 69.43 & 59.30 & \textbf{\underline{78.26}} & \textbf{68.31} & \textbf{\underline{74.26}} & 70.71 & 70.04 \\
\textbf{ElhBERTeu}$^{\dagger}$ & 110M & \textbf{\underline{69.89}} & \textbf{\underline{63.81}} & 78.05 & 65.93 & 73.84 & \textbf{\underline{71.71}} & \textbf{\underline{70.54}} \\
\midrule
 \textbf{GPT-3.5 Turbo} & n/a & 59.52 & 38.17 & 42.66 & 50.09 & 50.00 & 51.29 & 48.62 \\
 \textbf{GPT-4 Turbo}   & n/a & \textbf{67.90} & \textbf{57.10} & \textbf{52.21} & \textbf{\underline{88.25}} & \textbf{71.85} & \textbf{61.93} & \textbf{66.54} \\
\midrule
\textbf{XGLM}      & 7B & 39.94 & 21.58 & 36.73 & 50.94 & 50.42 & 49.21 & 41.47 \\
 \textbf{BLOOM}    & 7B & 37.94 & 20.72 & 39.10 & 48.21 & 47.48 & 47.57 & 40.17 \\
 \textbf{Mistral}  & 7B & 40.63 & 21.40 & 24.81 & 48.04 & 50.84 & 49.57 & 39.22 \\
  \textbf{Llama 2} & 7B & 41.63 & 18.60 & 20.06 & 50.94 & 48.32 & 49.64 & 38.20\\
 \rowcolor{purple!5} \new{}  \textbf{Latxa}   & 7B & \textbf{57.30} & \textbf{45.65} & \textbf{57.44} & \textbf{49.50} & \textbf{54.20} & \textbf{51.28} & \textbf{52.56} \\
\midrule
  \textbf{mGPT}    & 13B & 35.41 & 17.54 & 23.73 & 47.53 & 50.84 & 50.29 & 37.56 \\
  \textbf{Llama 2} & 13B & 41.09 & 18.25 & 27.35 & 49.23 & 48.74 & 49.21 & 38.98\\
 \rowcolor{purple!5} \new{}  \textbf{Latxa}   & 13B & \textbf{53.92} & \textbf{47.66} & \textbf{57.50} & \textbf{54.17} & \textbf{55.88} & \textbf{51.00} & \textbf{53.36} \\
  \midrule
  \textbf{Mixtral} & 8x7B & 49.46 & 21.81 & 37.32 & 53.32 & 57.56 & 50.50 & 45.00 \\
  \textbf{Yi}       & 34B & 47.08 & 29.33 & 30.69 & 54.68 & 49.58 & 52.00 & 43.90 \\
  \textbf{Llama 2}  & 70B & 47.47 & 21.01 & 31.01 & 52.98 & 51.26 & 51.57 & 42.55\\ 
 \rowcolor{purple!5} \new{}  \textbf{Latxa}    & 70B & \textbf{61.06} & \textbf{55.71} & \textbf{55.88} & \textbf{72.57} & \textbf{59.66} & \textbf{53.57} & \textbf{59.74} \\
 \bottomrule
\end{NiceTabular}
\caption{BasqueGLUE results by task. *VaxxStance is measured in terms of macro-average F1-score of the categories \textsc{in favour} and \textsc{against}. $^{\dagger}$BERTeus and ElhBERTeu are fine-tuned encoders.}
\label{tab:basqueglue}
\end{table*}

\paragraph{Effectiveness of continued pretraining.} Latxa obtains the best results in each compute class, outperforming all previous open models by a large margin. As the only exception, Mistral 7B is better than Latxa 7B on Belebele and EusReading, but Latxa 7B wins in all the other 5 datasets. Our best model obtains 61.08 points on average, outperforming the previous best open model by 18.95 points. In addition, it outperforms the Llama 2 model it is based on by 25.18 points, confirming the effectiveness of continued pretraining to build language models for low-resource languages.

\paragraph{Open vs. closed models.} With the exception of EusProficiency, the best results are obtained by GPT-4 Turbo, a closed commercial system. The difference between GPT-4 Turbo and the previous best open model (Yi) is abysmal. For instance, GPT-4 Turbo is 30.55 points better than Yi on EusTrivia, while the latter is only 16.02 points better than random chance. This can partly be attributed to previous open initiatives being primarily English-centric. While Latxa substantially narrows this gap, we believe that future research on open models should pay more attention to low-resource languages.

\paragraph{English-centric vs. multilingual models.} The 3 multilingual models we evaluate (XGLM, BLOOM and mGPT) do better than all English-centric models on XStoryCloze, which requires linguistic competence in Basque and is evaluated in a zero-shot fashion. However, English-centric open models do generally better on other tasks, presumably due to their better in-context learning capabilities and general knowledge captured. Consistent with our previous point, this suggests that existing open models are either English-centric and struggle in low-resource languages like Basque, or are multilingual and significantly lag behind the English-centric models in language-agnostic capabilities.

\paragraph{Impact of scale.} We find that larger Latxa models obtain substantially better results: the 70B model is 10.99 points better than the 13B model on average, which is 9.14 points better than the 7B model. This transcends conventional scaling laws, which establish that, when pretraining models from scratch in low-resource scenarios, the performance is bottlenecked by the amount of training data rather than the model size \citep{kaplan2020scaling}. However, our results show a different picture for continued pretraining, where bigger and stronger base models result in better performance despite the limited pretraining data in the target language. This suggests that even better results could be obtained through continued pretraining as stronger English-centric models become available, which is encouraging for low-resource languages.

\paragraph{General vs. language-specific knowledge.} We find evidence that Latxa is particularly strong in tasks that test for proficiency in the Basque language. In particular, Latxa obtains the best results on Basque language proficiency exams (EusProficiency), despite lagging behind GPT-4 Turbo in the rest of the tasks. Similarly, Latxa outperforms GPT-4 Turbo on the \textit{Language \& Literature} subset of EusTrivia, even if GPT-4 Turbo is superior in the rest of the categories (\cref{tab:results-eustrivia}). This suggests that Latxa is more proficient than GPT-4 Turbo in Basque, but the latter does better in most tasks due to its stronger general capabilities. Another evidence of this is that Latxa is particularly weak in the \textit{Maths \& ICT} subset of EusTrivia, where it even lags behind Yi, while GPT-4 Turbo is particularly strong in this category. This is likely because most problems in this category can be understood with basic knowledge of Basque, but solving them may require more complex mathematical reasoning. This suggests that the general capabilities of language models are highly language-agnostic which, in line with our previous finding, suggests that stronger English-centric models can lead to stronger models for low-resource languages by using the same continued pretraining recipe.

\paragraph{Classical NLP tasks.} \cref{tab:basqueglue} reports fine-grained results on BasqueGLUE, which comprises various classical NLP tasks like topic classification and coreference detection. In addition to our usual set of decoder-only models evaluated in a few-shot fashion, we report results for BERTeus \citep{agerri-etal-2020-give} and ElhBERTeu \citep{urbizu-etal-2023-enough}, which are encoder-only models that were fine-tuned specifically on these tasks. The best results are obtained by these specialized encoder-only models, which shows that the traditional pretraining/fine-tuning paradigm with BERT-style models is still competitive for classical NLP tasks. The only exception is \textit{EpecKorrefBin}, where both GPT-4 Turbo and Latxa 70B perform substantially better than the fine-tuned encoder-only models. In future work, we would like to explore fine-tuning Latxa and other decoder-only models on these tasks.

\section{Related Work}

The amount of documents per language in Common Crawl\footnote{CC-MAIN-2024-22 crawl at \url{https://commoncrawl.github.io/cc-crawl-statistics/plots/languages}}  is  helpful to organize the literature. There is no agreed-upon definition of low-resource language, so we set five arbitrary buckets of languages following a logarithmic distribution: \emph{high} bucket with just English (rank 1, $46\%$ of latest crawl), \emph{high-medium} languages around Spanish (rank 5, $4.6\%$), \emph{medium} around Danish (rank 24, $0.46\%$), \emph{low} resource around Nepalese (rank 46, $0.044$), and \emph{very-low} around Somali (rank 81, $0.0046\%$). Basque would be low-resource (rank 52, $0.035\%$). 

While there were some early efforts to build \textbf{open multilingual language models} like XGLM~\citep{lin-etal-2022-shot}, BLOOM~\citep{workshop2023bloom}, mGPT \citep{shliazhko2022mgpt} and the translation oriented MADLAD-400 \citep{kudugunta2023madlad}, their performance significantly lags behind more recent English-centric models like Llama 2 \cite{touvron2023Llama} or Mistral \cite{jiang2023mistral}. Concurrent to our work, \citet{ustun2024aya} present Aya, a fine-tuned mT5 encoder-decoder \cite{xue-etal-2021-mt5}, which has been instruction-tuned and supports 101 languages.

In the case of \textbf{high-medium-resource} languages, different teams have focused on building models from scratch.  \citet{ekgren-etal-2022-lessons} built a Swedish 20B model which was favorably evaluated on perplexity, followed by \cite{ekgren-etal-2024-gpt-sw3}, which builds a 40B model for five Nordic languages, including low-resource Danish and Faroese, English and code. It was evaluated on perplexity for the three richest languages. \citet{faysse2024croissantllm} present a French-English bilingual model trained with the same number of tokens for each language, although the amount of text for French (376M documents) is half the English text. The authors stress the fact that the tokenizer should not be biased towards any of the two languages. The results on French (and English) LLM evaluation benchmarks show that their largest model (1.2B parameters) underperforms both 3B Llama 2 and Mistral models. 

Regarding \textbf{medium-resource},  \citet{luukkonen-etal-2023-fingpt} focus on Finnish. Using a 19B word corpus they trained several models from scratch, ranging from 186M to 13B parameters. As an alternative, they also \textbf{continued pretraining} a multilingual model, the 176B BLOOM. The evaluation on FIN-bench, a version of Big-Bench \cite{srivastava2023imitation}, shows the 13B model underperforms the 7B model, while the continued pretraining model obtains the best results by a large margin. In follow-up work, \citet{luukkonen2024poro} train a 40B bilingual model from scratch with a larger Finnish corpus, and obtain the best results in FIN-bench. They do not compare the results to commercial models. The results suggest that monolingual models trained from scratch saturate at relatively small sizes, and multilingual or continued pretraining would be the best option for these languages. 



Focusing on \textbf{very-low-resource} languages, \citet{yong-etal-2023-bloom} compare different approaches to extend the 176B BLOOM model from 46 languages to new languages unseen at training, including Guarani (rank 116, $0.0006\%$) and seven larger languages for which they sample 100K documents at most. Model sizes range from 0.56B to 7.1B parameters. They report good results for adapters on zero-shot benchmarks. 
In concurrent work, \citet{lin2024mala} present MaLA-500, a continued pretrained model using LORA based on Llama 2 that supports 500 languages. MaLA seems to improve Llama 2 on low to very-low resource languages, but degrades in some medium languages already covered in Llama 2. Evaluation is based on a variant of perplexity and text classification.

Previous work on \textbf{low-resource} languages has been done in the context of multilingual models (see above). In all cases, evaluation is based on perplexity and/or some readily available datasets, and does not include native benchmarks designed to evaluate base models.

\section{Conclusion and Future Work}

In this work, we present a new open framework for the development and evaluation of LLMs in Basque. The framework includes Latxa, a set of state-of-the-art generative LLMs that have been built by continuing to pretrain the Llama 2 7B, 13B and 70B models in Basque. The pretraining dataset is the largest public dataset available to date and includes data from carefully curated sources, as well as content derived from automatically filtered versions of Common Crawl. After preprocessing and deduplication, the released corpora comprise 1.22B words and 4.17B tokens. We also present 4 new evaluation datasets, collectively the largest evaluation benchmark for Basque
that allows assessing the knowledge of the models about the Basque language and culture. 

The Latxa models outperform all previous open models and GPT-3.5 Turbo, but they still lag behind GPT-4 Turbo in most benchmarks. Interestingly, Latxa 70B outperforms GPT-4 Turbo on EusProficiency and \textit{Language \& Literature} EusTrivia questions, suggesting that the capabilities of LLMs in a particular language are not determined by their linguistic competence in this language. This, along with the effectiveness of scale, suggests that applying the exact same continued pretraining recipe could lead to better Basque models as stronger English-only models become available.

In the future, we plan to extend the training dataset by gathering quality content from diverse Basque sources such as publishers or media, as well as building evaluation datasets to assess aspects such as truthfulness or hallucinations. We also plan to further tune Latxa to follow instructions, which should improve the overall capabilities of our models.




\section*{Limitations} \label{sec:limitations}

To alleviate the potentially disturbing or harmful content, Latxa has been trained on carefully selected and processed data which comes mainly from local media, national/regional newspapers, encyclopedias and blogs. Still, the model is based on Llama 2 models and can potentially carry the same biases, risks and limitations.

Latxa models are pretrained LLMs without any task-specific or instruction fine-tuning. The model was not fine-tuned to follow instructions or to work as a chat assistant, therefore, this kind of usage is not tested nor recommended. That is, the model can either be prompted to perform a specific task or further fine-tuned for specific use cases.

\section*{Acknowledgements}

This work has been partially supported by the Basque Government (Research group funding IT-1805-22 and IKER-GAITU project), the Spanish Ministry for Digital Transformation and of Civil Service, and the
EU-funded NextGenerationEU Recovery, Transformation and Resilience Plan
(ILENIA project, 2022/TL22/00215335).
The models were trained on the Leonardo supercomputer at CINECA under the EuroHPC Joint Undertaking, project EHPC-EXT-2023E01-013. Julen Etxaniz and Oscar Sainz hold a PhD grant from the Basque Government (PRE\_2023\_2\_0060 and PRE\_2023\_2\_0137, respectively).

\bibliography{custom, anthology}

\clearpage
\appendix

\section{Data Preprocessing Details}
\label{app:data}

\begin{figure*}
    \centering
    \subfigure[Cross-dataset document repetitions]{\includegraphics[width=0.5\textwidth,trim={0 10pt 0 25pt},clip]{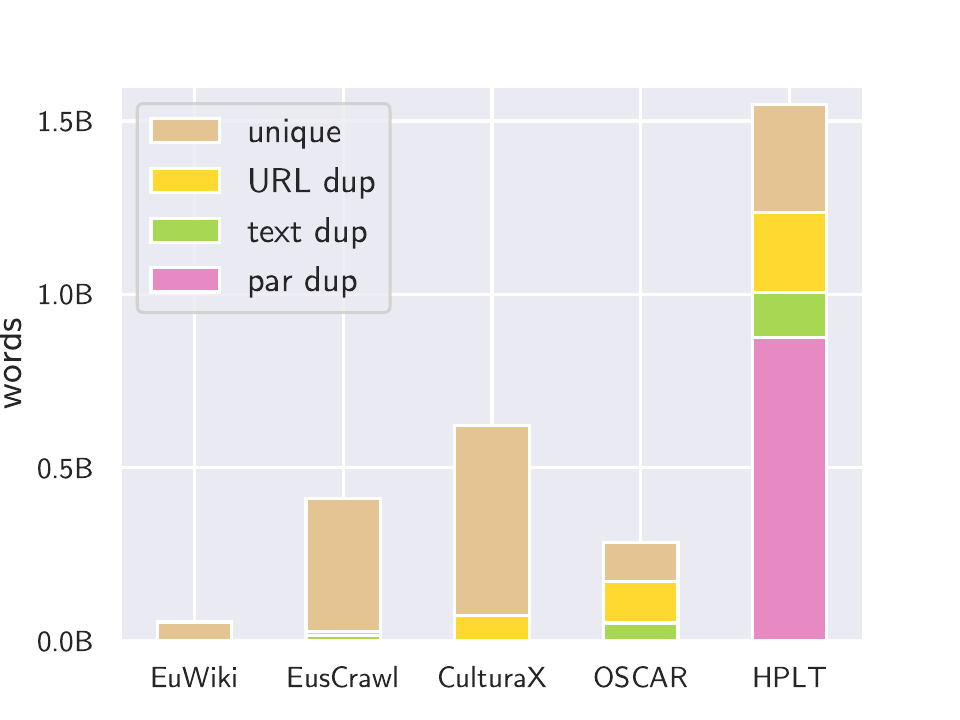}\label{fig:data-a}}\hfill
    \subfigure[\texttt{\#\;words:} Words per document]{\includegraphics[width=0.5\textwidth,trim={0 10pt 0 25pt},clip]{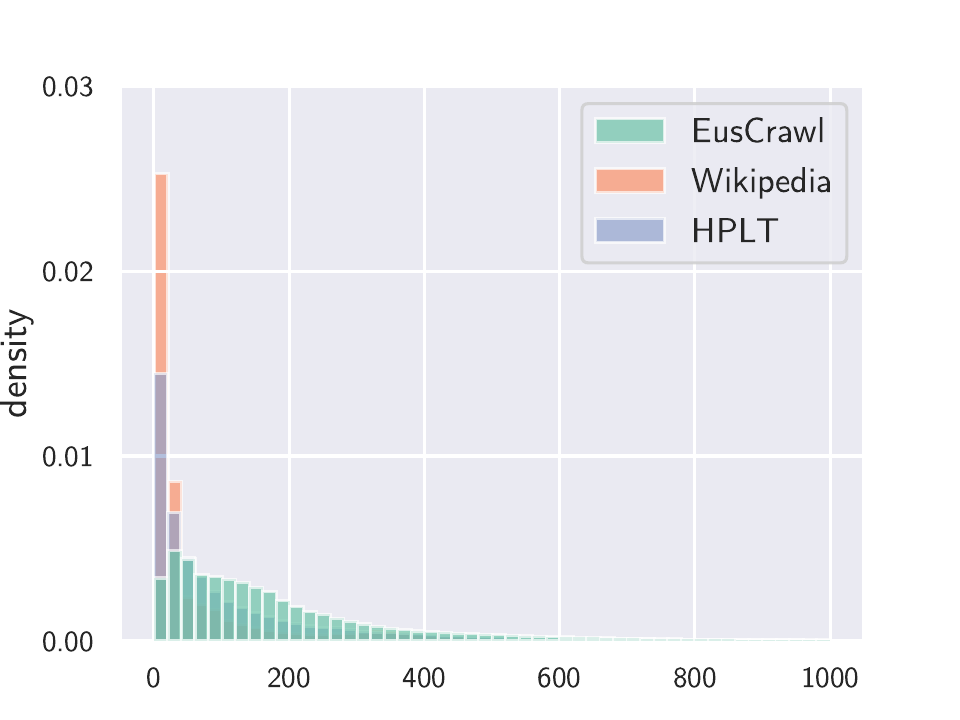}\label{fig:data-b}}
    \subfigure[\texttt{alpha}: alphabetic characters to word length ratio]{\includegraphics[width=0.5\textwidth,trim={0 10pt 0 25pt},clip]{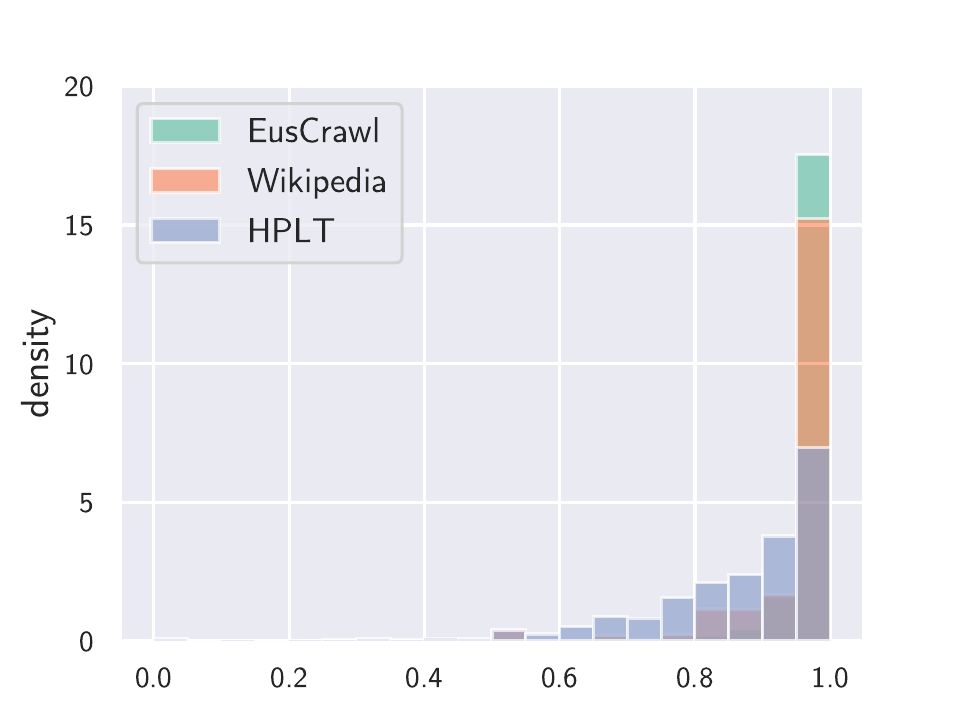}\label{fig:data-c}}\hfill
    \subfigure[\texttt{word\;len}: Medium characters per word]{\includegraphics[width=0.5\textwidth,trim={0 10pt 0 25pt},clip]{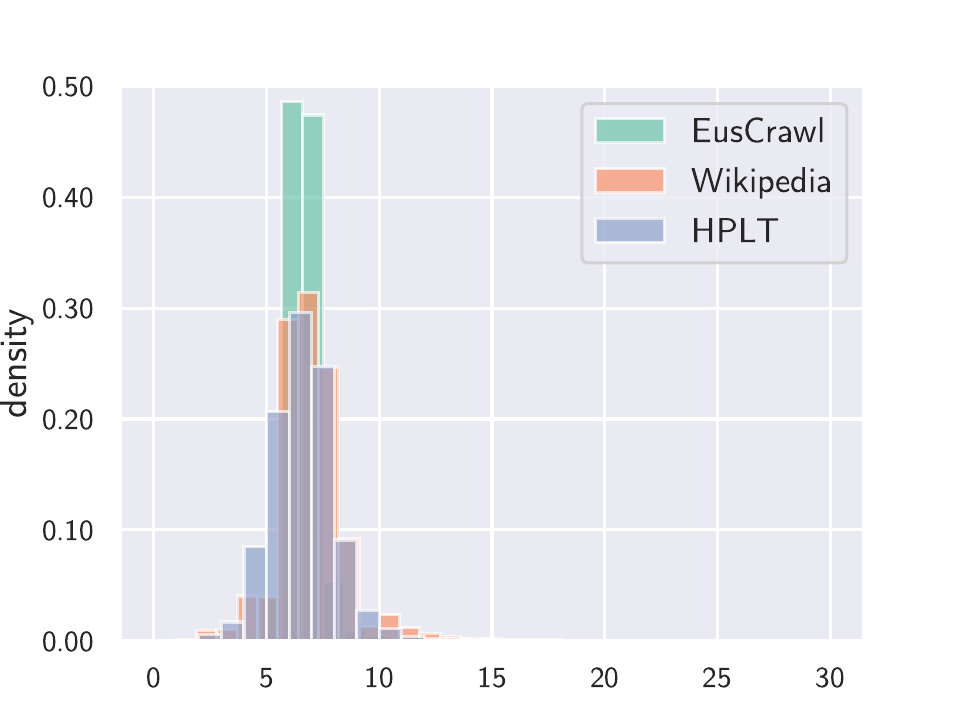}\label{fig:data-d}}
    \caption{Basic corpus quality statistics before preprocessing}
    \label{fig:data}
\end{figure*}

The mix of raw documents ---7.39 million documents, 2.96 billion words--- was deduplicated with Dolma \citep{soldaini2024dolma} to discard both intra- and cross-dataset document repetitions. The number of words discarded per dataset at this stage is illustrated in \cref{fig:data-a}, where datasets are shown in order of preference (Egunkaria and Booktegi are omitted due to their size). In the case of EusCrawl v1.1, the duplicate documents arise mainly from an overlap between EusCrawl v1 and the updated content. For Colossal OSCAR, more than 60\% of the documents are already present in the preferred datasets (i.e., EusCrawl, Egunkaria, Booktegi, EuWiki or CulturaX). HPLT was further deduplicated at the paragraph level due to its quality. After deduplication, the pretraining corpus amounts to 5.80 million documents and 1.45 billion words.

The deduplicated corpus was further filtered based on document-level features. Specifically, we applied Dolma's implementation of a set of heuristics from Gopher \cite{rae2022scaling} and C4 \cite{raffel2020exploring}, and the Corpus Cleaner v2 \citep[CCv2;][]{palomar-giner-etal-2024-curated-catalog}. The document-level features are as follows:

\begin{itemize}[noitemsep]
    \item \texttt{\small eu}: percentage of the text that is written in Basque according to CLD2
    \item \texttt{\small \#\;words}: number of words
    \item \texttt{\small word\;len}: mean word length
    \item \texttt{\small bullet}: fraction of lines starting with `*' or `-'
    \item \texttt{\small ellipsis}: fraction of lines ending with `…'
    \item \texttt{\small lorem\;ipsum}: whether it occurs in the text
    \item \texttt{\small brackets}: whether `\{' occurs in the text
    \item \texttt{\small symbols}: `\#' and  `…' to word ratio
    \item \texttt{\small alpha}: fraction of words with at least one alphabetic character
    \item \texttt{\small CCv2}: aggregated score given by CCv2
\end{itemize}

\noindent The threshold applied to each feature and its impact on the datasets is shown in \cref{tab:data-tags}. We adapted Dolma's default thresholds regarding document and word length to better fit our data and the Basque language based on observed distributions of our corpora (see examples in \cref{fig:data}). Note that the filters are not mutually exclusive, that is, the same document might be flagged for removal by several filters. 

As a result of this process, the least curated corpora, HPLT and Colossal OSCAR, were further reduced by 61\% and 5\% respectively, in terms of words. The 22.37\% EuWiki documents flagged as being too short correspond to Wikipedia's redirect pages, which ultimately did not affect the word count significantly. The final size of the pretraining corpus is 4.30 million documents and 1.22 billion words, which equates to 4.17B Llama 2 tokens.

\setcmapmax{100}
\begin{table*}
    \centering
    \small
    \begin{tabular}{l*{8}{N}}
    \toprule
                              & \textbf{CulturaX} & \textbf{EusCrawl} & \textbf{~~HPLT~~}  & \textbf{OSCAR} & \textbf{EuWiki} & \textbf{Egunkaria} & \textbf{Booktegi} \\ 
    \midrule
    \texttt{eu} \textless~0.5          & 0.00     & 5.30    & 34.13 & 0.00  & 0.00      & 0.00      & 0.00     \\
    \texttt{\#\;words} \textless~4     & 0.00     & 0.04     & 3.61  & 0.13  & 22.37     & 0.06      & 0.00     \\
    \texttt{word\;len} \textless~3     & 0.00     & 0.04     & 0.66  & 0.02  & 0.87      & 0.11      & 0.00     \\
    \texttt{word\;len} \textgreater~12 & 0.00     & 1.12     & 0.88  & 0.36  & 1.26      & 0.00      & 0.00     \\
    \texttt{alpha} \textless~0.8       & 0.11     & 1.21     & 23.93 & 19.14 & 4.19      & 0.84      & 0.00     \\
    \texttt{symbols} \textgreater~0.1  & 0.09     & 0.05     & 0.47  & 0.23  & 0.02      & 0.00      & 0.00     \\
    \texttt{ellipsis} \textgreater~0.3 & 0.74     & 0.09     & 0.25  & 3.47  & 0.00      & 0.01      & 0.00     \\
    \texttt{bullets} \textgreater~0.9  & 0.03     & 0.01     & 0.01  & 0.17  & 0.00      & 0.00      & 0.00     \\   
    \texttt{lorem\;ipsum}              & 0.00     & 0.00     & 0.02  & 0.31  & 0.00      & 0.00      & 0.00     \\
    \texttt{brackets}                  & 0.43     & 0.07     & 0.73  & 16.30 & 0.23      & 0.01      & 1.13     \\
    \midrule
    \texttt{CCv2} \textless~0.55      & 0.04 & 0.56 & 25.12 & 0.11 & 1.62 & 1.19 & 5.08 \\
    \bottomrule
    \end{tabular}
    \caption{Percentage of documents dropped by each filter.}
    \label{tab:data-tags}
\end{table*}

\section{Dataset Contamination}
\label{app:contamination}

Following previous work on data contamination \cite{brown2020language,touvron2023Llama}, we check for token n-gram overlaps between test items and training data. To that end, we index training documents in Elasticsearch, applying the standard tokenizer to lowercase text, and removing stopwords (built-in Basque stopwords and all auxiliary verbs).

Given that test items vary greatly in length from one benchmark to another, we avoid establishing an arbitrary n-gram length threshold above which to consider a test item contaminated. Instead, we report statistics based on the longest n-grams matched, spanning our search from n-grams equal to each item's total length to just one word. For each n-gram size $n$ we only considered test items of equal or bigger size when assessing contamination. 

Results are summarized in \cref{tab:contamination}, which reports the contamination percentages (\textit{cont \%}) across each benchmark with respect to a specific quartile of test questions or context lengths (\textit{n}). Higher contamination values are to be expected at lower quartiles, as shorter n-grams (typically 1 to 5 words) tend to involve frequent word combinations and are thus more likely to overlap with the training data. We observe that contamination percentages tend to decrease to near zero after the first quartile, with the following exceptions. 

Notably, QNLI has substantial overlap even at higher n-gram lengths. This is explained by the fact that QNLI contexts were taken from Wikipedia. In line with the analysis of \citet{palm2023aakanksha}, we do not consider these items to be contaminated, as the questions and answers have not been found alongside the contexts in the training data. 

In the case of EusProficiency, it must be noted it comprises particularly short test items, the median length of a question being 4 words. Upon manual analysis, we did not observe any annotation contamination.

As for EusExams, this evaluation benchmark consists of Public Service examinations and thus contains many references to national and regional laws, directives and plans, services, administration offices, etc. Such mentions can also be found on the web, as this information is of public interest. Again, we did not observe annotation contamination.


\begin{table*}
    \small
    \centering
    \begin{tabular}{lrrrrrrrrrr}
    \toprule
     & \multicolumn{2}{c}{\textbf{min}} & \multicolumn{2}{c}{\textbf{25\%}} & \multicolumn{2}{c}{\textbf{50\%}} & \multicolumn{2}{c}{\textbf{75\%}} & \multicolumn{2}{c}{\textbf{max}} \\
     \cmidrule(lr){2-3} \cmidrule(lr){4-5} \cmidrule(lr){6-7} \cmidrule(lr){8-9} \cmidrule(lr){10-11}
     & \textbf{n} & \textbf{cont \%} & \textbf{n} & \textbf{cont \%} & \textbf{n} & \textbf{cont \%} & \textbf{n} & \textbf{cont \%} & \textbf{n} & \textbf{cont \%} \\
    \midrule
    \textbf{Belebele} & 18 & 0.0 & 40 & 0.0 & 51 & 0.0 & 64 & 0.0 & 138 & 0.0 \\
    \textbf{XStory Cloze} & 1 & 100.0 & 4 & 17.4 & 5 & 1.3 & 6 & 0.1 & 12 & 0.0 \\
    \midrule
    \textbf{EusProficiency} & 1 & 99.7 & 3 & 34.1 & 4 & 6.4 & 6 & 0.6 & 21 & 0.0 \\
    \textbf{EusReading} & 1 & 100.0 & 5 & 88.1 & 57 & 0.0 & 578 & 0.0 & 808 & 0.0 \\
    \textbf{EusTrivia} & 1 & 100.0 & 4 & 7.1 & 5 & 1.3 & 7 & 0.0 & 21 & 0.0 \\
    \textbf{EusExams} & 1 & 96.6 & 6 & 13.6 & 9 & 7.1 & 15 & 0.1 & 105 & 0.0 \\
    \midrule
    \textbf{BEC} & 1 & 100.0 & 8 & 0.2 & 11 & 0.0 & 13 & 0.0 & 34 & 0.0 \\
    \textbf{BHTCv1} & 1 & 99.8 & 19 & 2.8 & 25 & 0.4 & 34 & 0.1 & 147 & 0.0 \\
    \textbf{Korref} & 8 & 0.0 & 21 & 0.0 & 25 & 0.0 & 33 & 0.0 & 66 & 0.0 \\
    \textbf{QNLI} & 1 & 100.0 & 3 & 97.1 & 5 & 71.0 & 10 & 16.4 & 84 & 0.0 \\
    \textbf{VaxxStance} & 2 & 99.4 & 15 & 0.3 & 22 & 0.0 & 29 & 0.0 & 39 & 0.0 \\
    \textbf{WiC} & 2 & 100.0 & 12 & 2.6 & 18 & 0.4 & 26 & 0.0 & 42 & 0.0 \\
    \bottomrule
    \end{tabular}
    \caption{Data contamination results for all our evaluation datasets. The table shows the contamination percentage (\textit{cont \%}) considering different n-gram sizes (\textit{n}) that depend on the length of each dataset's items.}
    \label{tab:contamination}
\end{table*}

\section{Task Examples and Prompts}
\label{app:examples_prompts}

Tables \ref{tab:eval-framework-eus} and \ref{tab:eval-framework-basqueglue} contain, respectively, the prompt templates and examples of our new datasets (i.e., EusProficiency, EusReading, EusTrivia, and EusExams) and BasqueGLUE tasks. ``\textit{System}'' refers to system prompts and applies only to GPT evaluations. Additionally, the number of shots and metrics used to measure the results are also specified per task. 

\begin{table*}[t]
    \small
    \centering
    \begin{NiceTabularX}{\textwidth}{rX}
        \toprule
        \multicolumn{2}{l}{\textbf{EusProficiency}} \textit{5-shot, accuracy} \\
        \cmidrule(lr){1-1}
         System  & Respond always with a single letter: A, B, C or D.\\
         Prompt  & \verb|Galdera: {question}\nA. {opt[0]}\nB. {opt[1]} ... \nErantzuna:|\\
         Example & Galdera: Jatetxe batera sartu, eta bazkaltzen ari denari: \\
                 & A. Gabon! \\
                 & B. On egin diezazula! \\
                 & C. Bejondeizula! \\
                 & D. Agur t'erdi! \\ 
                 & Erantzuna: B \\
                 \cmidrule(lr){2-2}
                 & {\scriptsize \textit{Question: Upon entering a restaurant, to another diner: A. Good night! B. Enjoy! C. Bless you! D. Greetings! Answer: B}} \\
        \midrule
        \multicolumn{2}{l}{\textbf{EusReading}} \textit{1-shot, accuracy} \\
        \cmidrule(lr){1-1}
         System  & Respond always with a single letter: A, B, C or D.\\
         Prompt  & \verb|Pasartea: {context}\n\nGaldera: {question}\nA. {opt[0]}\nB. {opt[1]} ... \nErantzuna:|\\
         Example & Pasartea: Ernest Hemingway, berak jakin barik, azkenekoz etorri da Bilbora, eta oro har, Penintsulara. Eta hori tamala, hilak 24 dituelarik Bilbon zezenak Ordoñez bere kutunari adarkada ederra sartu dio. Ez da ezer izan, zorionez. Biharamuneko El Correo Español egunkarian emandako argazkian ikusten den legez, idazleak bisita egin dion unean, toreatzailea hortxe dago, ondo bizirik, ohean. [...] \\
                 & \\
                 & Galdera: 1960ko abuztuaren 24an \\
                 & A. El Correo Español-eko C. Barrenarekin batera agertzen den Ordoñez toreatzailea harrapatu zuen zezen batek. \\
                 & B. Ernest Hemingwayk Bilboko plazan adarkada jaso zuen Ordoñez toreatzaileari bisita egin zion. \\
                 & C. El Correo-ko argazkian, zezen batek Ordoñez toreatzailea harrapatzen du. \\
                 & D. Ernest Hemingway lehenengo eta azkeneko aldiz iritsi zen Bilbora.\\
                 & Erantzuna: B\\
                 \cmidrule(lr){2-2}
                 & {\scriptsize \textit{Passage: Ernest Hemingway, without his knowledge, came for the last time to Bilbao, and in general to the Peninsula. And indeed, on the 24th, at Bilbao, the bull gave his favourite Ordóñez a good goring. It was nothing, luckily. As can be seen from the photograph published in El Correo Español the next day, the bullfighter is there, alive, in bed, the moment the writer visits him. [...] Question: On August 24, 1960 A. The bullfighter Ordóñez, who appears next to C. Barrena of El Correo Español, was caught by a bull. B. Ernest Hemingway visited the bullfighter Ordóñez, who had received a goring in the square of Bilbao. C. In the photo of El Correo, a bull catches the bullfighter Ordóñez. D. Ernest Hemingway arrived in Bilbao for the first and last time. Answer: B}} \\
        \midrule
        \multicolumn{2}{l}{\textbf{EusTrivia}} \textit{5-shot, accuracy} \\
        \cmidrule(lr){1-1}
         System  & Respond always with a single letter: A, B, C or D.\\
         Prompt  & \verb|Galdera: {question}\nA. {opt[0]}\nB. {opt[1]} ... \nErantzuna:|\\
         Example & Galdera: Zenbat kilo dauka tona batek? \\
                 & A. 10.000 kilo \\
                 & B. 1.000.000 kilo \\
                 & C. 1.000 kilo \\
                 & D. 100 kilo \\
                 & Erantzuna: C \\
                 \cmidrule(lr){2-2}
                 & {\scriptsize \textit{Question: How many kilograms are there in a tonne? A. 10,000 kilos B. 1,000,000 kilos C. 1,000 kilos D. 100 kilos Answer: C}} \\
        \midrule
        \multicolumn{2}{l}{\textbf{EusExams}} \textit{5-shot, accuracy} \\
        \cmidrule(lr){1-1}
         System  & Respond always with a single letter: A, B, C or D.\\
         Prompt  & \verb|Galdera: {question}\nA. {opt[0]}\nB. {opt[1]} ... \nErantzuna:|\\
         Example & Galdera: UPV/EHUREN ONDAREA HAU DA: \\
                  & A. UPV/EHUk jabetzan dituen ondasunak. \\
                  & B. UPV/EHUk jabetzan dituen ondasun eta eskubideak. \\
                  & C. UPV/EHUk jabetzan edo titularitatean dituen ondasun eta eskubideak, bai eta etorkizunean eskuratzen edo esleitzen zaizkion gainerako guztiak ere. \\
                  & D. UPV/EHUk jabetzan dituen ondasunak, bai eta etorkizunean eskuratzen dituen gainerako guztiak ere. \\
                  & Erantzuna: C \\
                  \cmidrule(lr){2-2}
                  & {\scriptsize \textit{Question: UPV/EHU'S LEGACY IS: A. The property owned by UPV/EHU. B. The rights and property owned by the UPV/EHU. C. The rights and property of the UPV/EHU in ownership, as well as any other property acquired or assigned to it in the future. D. The property of the UPV/EHU in ownership, as well as any other property acquired or assigned to it in the future. Answer: C}} \\
        \bottomrule
    \end{NiceTabularX}
    \caption{Prompt templates and examples of the new evaluation tasks. ``\textit{System}'' refers to system prompts and only applies to GPT evaluations. Translations of the examples are given in \textit{italics}.}
    \label{tab:eval-framework-eus}
\end{table*}

\begin{table*}[t]
    \small
    \centering
    \begin{NiceTabularX}{\textwidth}{rX}
        \toprule
        \multicolumn{2}{l}{\textbf{BEC}} \textit{5-shot, micro F1} \\
        \cmidrule(lr){1-1}
         System  & Respond always with one of these: negatiboa, neutrala, positiboa \\
         Prompt  & \verb|Testua: {context}\nGaldera: Nolako jarrera agertzen du aurreko testuak?\nErantzuna:|\\
         Example & Testua: Eta Euskal Herrian noizko @eajpnv ? \#URL \\
                 & Galdera: Nolako jarrera agertzen du aurreko testuak? \\
                 & Erantzuna: negatiboa \\
                 \cmidrule(lr){2-2}
                 & {\scriptsize \textit{Text: And in the Basque Country when @eajpnv ? \#URL Question: What sentiment does the previous text convey? Answer: negative}} \\
        \midrule
        \multicolumn{2}{l}{\textbf{VaxxStance}} \textit{5-shot, macro F1 (\textsc{n favour} \& \textsc{against})} \\
        \cmidrule(lr){1-1}
         System  & Respond always with one of these: bai, ez \\
         Prompt  & \verb|Testua: {context}\nGaldera: Nolako jarrera agertzen du aurreko testuak txertoei buruz?\nErantzuna:|\\
         Example & Testua: 45 urtetik gorakoen txertaketa hasiko da igandean Israelenhttps://t.co/6opid1ULyd \\
                 & Galdera: Nolako jarrera agertzen du aurreko testuak txertoei buruz? \\
                 & Erantzuna: neutrala \\
                 \cmidrule(lr){2-2}
                 & {\scriptsize \textit{Text: vaccination of over 45 years of age begins on Sunday in Israelhttps://t.co/6opid1ULyd Question: What stance does the previous text take on vaccines? Answer: neutral}} \\
        \midrule
        \multicolumn{2}{l}{\textbf{BHTCv2}} \textit{5-shot, micro F1} \\
        \cmidrule(lr){1-1}
         System  & Respond always with one of these: Ekonomia, Euskal Herria, Euskara, Gizartea, Historia, Ingurumena, Iritzia, Komunikazioa, Kultura, Nazioartea, Politika, Zientzia \\
         Prompt  & \verb|Testua: {context}\nGaldera: Zein da aurreko testuaren gaia?\nErantzuna:|\\
         Example & Testua: Eusko Jaurlaritza ari da prestatzen eta EAEko ikastetxe guztietara zabaltzeko asmoa du. \\
                 & Galdera: Zein da aurreko testuaren gaia? \\
                 & Erantzuna: Gizartea \\
                 \cmidrule(lr){2-2}
                 & {\scriptsize \textit{Text: The Basque Government is preparing it and intends to extend it to all schools in the Basque Country. Question: What is the subject of the previous text? Answer: Society}} \\
        \midrule
        \multicolumn{2}{l}{\textbf{EpecKorrefBin}} \textit{5-shot, accuracy} \\
        \cmidrule(lr){1-1}
         System  & Respond always with one of these: ez, bai \\
         Prompt  & \verb|Testua: {context}\nGaldera: Aurreko testuan, "*{w1}*" eta "*{w2}*" gauza bera dira?\nErantzuna:|\\
         Example & Testua: *Luis Uranga* harrituta azaldu da Portugalgo klubaren jokaerarekin. RICARDO SA PINTOK datorren denboraldian Lisboako Sportingen jokatu zuela pentsatzen genuen guztiok, baina une honetan Lisboarako bidea erabat zaildu zaio *Realeko aurrelariari*. \\
                 & Galdera: Aurreko testuan, "*Luis Uranga*" eta "*Realeko aurrelariari*" gauza bera dira? \\
                 & Erantzuna: ez \\
                 \cmidrule(lr){2-2}
                 & {\scriptsize \textit{Question: *Luis Uranga* was surprised by the way the Portuguese club acted. We all thought RICARDO SA PINTO would be playing next season at the Sporting of Lisbon, but right now the road to Lisbon has become very difficult for *the forward of La Real*. In the previous text, is "*Luis Uranga*" the same as "*Realeko aurrelariari*"? Answer: no}} \\
        \midrule
        \multicolumn{2}{l}{\textbf{QNLI$_{eu}$}} \textit{5-shot, accuracy} \\
        \cmidrule(lr){1-1}
         System  & Respond always with one of these: bai, ez \\
         Prompt  & \verb|{question}\n{context}\nGaldera: aurreko galderari erantzuten al dio emandako testuak?\nErantzuna:|\\
         Example & Nortzuen lehen alaba izan zen Dua Lipa? \\
                 & Liparen lehen hezkuntzako ikasketak musika klaseak eduki zituen, eta jotzen ikasi zuen lehen instrumentua biolontxeloa izan zen. \\
                 & Galdera: aurreko galderari erantzuten al dio emandako testuak? \\
                 & Erantzuna: ez \\
                 \cmidrule(lr){2-2}
                 & {\scriptsize \textit{Whose first daugther was Dua Lipa? Lipa's primary education included music lessons, and the first instrument she learned to play was the cello. Question: does the text given answer the previous question? Answer: No.}} \\
        \midrule
        \multicolumn{2}{l}{\textbf{WiC$_{eu}$}} \textit{5-shot, accuracy} \\
        \cmidrule(lr){1-1}
         System  & Respond always with one of these: bai, ez\\
         Prompt  & \verb|1. esaldia: {sent[0]}\n2. esaldia: {sent[1]}\nGaldera: Aurreko bi esaldietan, "{word}" hitzak|\\
                 & \verb|esanahi berdina du?\nErantzuna:| \\
         Example & 1. esaldia: beste alde batetik, irakasleek materiala prestatzeko dituzten aukera informatikoak ere gero eta ugariagoak dira; \\
                 & 2. esaldia: Unitate horretan konturatuko zinen bezala, materialak aldakorrak dira: batzuk lurrindu egiten dira berotzen direnean, beste batzuk apurtu edo eraldatu, edo aldaketa kimikoak jasan ditzakete. \\
                 & Galdera: Aurreko bi esaldietan, "material" hitzak esanahi berdina du? \\
                 & Erantzuna: ez \\
                 \cmidrule(lr){2-2}
                 & {\scriptsize \textit{Sentence 1: on the other hand, the computer possibilities for teachers to prepare materials are increasing; Sentence 2: As you may have noticed in that unit, materials are changeable: some evaporate when heated, others brake or transform, or they may undergo chemical changes. Question: In the two previous sentences, does the word "material" have the same meaning? Answer: no}} \\
        \bottomrule
    \end{NiceTabularX}
    \caption{Prompt templates and examples of BasqueGLUE tasks. ``\textit{System}'' refers to system prompts and only applies to GPT evaluations. Translations of the examples are given in \textit{italics}.}
    \label{tab:eval-framework-basqueglue}
\end{table*}

\section{Model Card}
\label{app:model-card}

We report Latxa's model card in \cref{tab:model-card}.

\begin{table*}[t]
    \small
    \centering
    \begin{NiceTabularX}{\textwidth}{lX}
        \toprule
        \multicolumn{2}{l}{\textbf{Model Details}} \\
        \cmidrule(lr){1-1}
        \textit{Model Developers} & (Anonymous upon publication) \\
        \textit{Variations} & Latxa comes in a range of parameter sizes: 7B, 13B, and 70B. \\
        \textit{Input} & Models input text only. \\
        \textit{Output} & Models generate text only. \\
        \textit{Model Architecture} & Latxa, similar to Llama 2, is an auto-regressive language model that uses an optimized transformer architecture. \\
        \textit{Model Dates} & Latxa was trained between October 2023 and February 2024. \\
        \textit{Status} & This is a static model trained on an offline dataset. Future versions of the model may include more updated data. \\
        \textit{License} & Latxa is based on Llama 2 models, and therefore, inherits their license. It is a custom commercial license available at: \url{https://ai.meta.com/resources/models-and-libraries/llama-downloads/} \\
        \textit{Where to send comments} & (Anonymous upon publication) \\ \midrule

        \multicolumn{2}{l}{\textbf{Intended Use}} \\
        \cmidrule(lr){1-1}
        \textit{Intended Use Cases} & Latxa models are intended to be used with Basque data; for any other language, the performance is not guaranteed. Latxa inherits the Llama 2 License which allows for commercial and research use. Latxa family models are pretrained LLMs without any task-specific or instruction fine-tuning. That is, the model can either be prompted to perform a specific task or further fine-tuned for specific use cases. \\
        \textit{Out-of-Scope Uses} & The model was not fine-tuned to follow instructions or to work as a chat assistant, therefore, this kind of usage is not tested nor recommended. \\ \midrule

        \multicolumn{2}{l}{\textbf{Hardware and Software} (Section~\ref{sec:models})} \\
        \cmidrule(lr){1-1}
        \textit{Training Factors} & The training of Latxa was conducted using GPT-Neox library. As infrastructure, we leveraged the CINECA HPC Leonardo computing cluster located in Italy. At most, 256 custom A100 GPUs were used to train the models. \\
        \textit{Carbon Footprint} & Pretraining utilized a cumulative 34.7K GPU hours of computation on hardware of type A100 64Gb (TDP 440W). Estimated total emissions were 4.53tCO\textsubscript{2}eq. \\ \midrule

        \multicolumn{2}{l}{\textbf{Training Data} (Section~\ref{sec:corpora})} \\
        \cmidrule(lr){1-1}
        \textit{Overview} & Latxa is trained on corpora from different sources. In general, quality was preferred over quantity, but content derived from automatically filtered versions of CommonCrawl was also included. After collecting the corpora, it was cleaned and deduplicated. Pretraining corpora includes: EusCrawl v1.1, Egunkaria, Booktegi, EuWiki, CulturaX, Colossal OSCAR, and, HLPT v1. \\
        \textit{Data Freshness} & The pretraining data has a cutoff of November 2023.\\ \midrule

        \multicolumn{2}{l}{\textbf{Evaluation}} \\
        \cmidrule(lr){1-1}
        & See Evaluation Data (Section~\ref{sec:new-datasets}), Experimental Setting (Section~\ref{sec:experimental-setting}), and Results (Section~\ref{sec:results}) \\ \midrule

        \multicolumn{2}{l}{\textbf{Ethical Considerations and Limitations} (Section~\ref{sec:limitations})} \\
        \cmidrule(lr){1-1}
        & To alleviate the potentially disturbing or harmful content, Latxa has been trained on carefully selected and processed data which comes mainly from local media, national/regional newspapers, encyclopedias and blogs. Still, the model is based on Llama 2 models and can potentially carry the same biases, risks and limitations. \\
        \bottomrule
    \end{NiceTabularX}
    \caption{Model card for Latxa}
    \label{tab:model-card}
\end{table*}

\section{Detailed EusExams Results}
\label{app:detailed-results}


We provide detailed results for EusExams by category in \cref{tab:results-eusexams}. Results are consistent across categories, our models outperform every model in the same size category by a large margin. Latxa 13B outperforms GPT-3.5 Turbo in most categories, but Latxa 70B is far from GPT-4 Turbo performance. This is expected as all categories in these exams require advanced knowledge. Health System is the most challenging category, followed by City Council. Public Office and University tests are easier for most models. For specific results of each test check \cref{tab:results-eusexams-detailed}.

\begin{table*}[ht]
\centering
\small
\begin{adjustbox}{max width=\linewidth}
\begin{tabular}{rr c c c c c}
\toprule
 &  & \textbf{Public Office} & \textbf{University} & \textbf{City Council} & \textbf{Health System} & \textbf{Average} \\ \midrule
\textbf{GPT-3.5 Turbo} & n/a & 47.29 & 43.43 & 41.61& 40.50 & 42.42\\
\textbf{GPT-4 Turbo} & n/a & \underline{\textbf{76.64}} & \underline{\textbf{76.22}} & \underline{\textbf{69.63}} & \underline{\textbf{64.61}} & \underline{\textbf{70.22}}\\ \midrule
\textbf{XGLM} & 7B & 23.82 & 23.38 & 24.82& 25.88 & 24.75\\
\textbf{BLOOM} & 7B & 24.08 & 24.02 & 25.56& 24.92 & 24.57\\
\textbf{Mistral} & 7B & 34.72 & 33.38 & 27.84& 28.37 & 30.76\\
\textbf{Llama 2} & 7B & 30.35 & 27.68 & 31.98& 28.26 & 28.63\\
\textbf{Latxa} & 7B & \textbf{37.22} & \textbf{37.29} & \textbf{35.24} & \textbf{29.39} & \textbf{33.32} \\ \midrule
\textbf{mGPT} & 13B & 25.01 & 24.47 & 24.80& 28.09 & 26.31\\
\textbf{Llama 2} & 13B & 31.99 & 30.52 & 26.81& 26.52 & 28.54\\
\textbf{Latxa} & 13B & \textbf{48.78} & \textbf{50.35} & \textbf{45.52} & \textbf{36.66} & \textbf{43.19}\\ \midrule
\textbf{Mixtral} & 8x7B & 44.87 & 42.73 & 38.47& 35.54 & 39.26\\
\textbf{Yi} & 34B & 41.13 & 41.76 & 36.61& 36.18 & 38.62\\
\textbf{Llama 2} & 70B & 37.42 & 34.75 & 30.78& 27.94 & 31.55\\
\textbf{Latxa} & 70B & \textbf{58.16} &\textbf{56.26} & \textbf{50.04} & \textbf{43.78} & \textbf{50.07}\\
\bottomrule
\end{tabular}
\end{adjustbox}
\caption{Detailed accuracy results over EusExams categories. Best results in each compute class are in \textbf{bold}. Best overall results are \underline{underlined}.}
\label{tab:results-eusexams}
\end{table*}

\begin{table*}[ht]
\centering
\small
\begin{adjustbox}{max width=\linewidth}
\begin{tabular}{r | c  c | c  c  c  c c | c  c  c | c  c  c  c }
\toprule
 & \textbf{GPT-3.5} & \textbf{GPT-4} & \textbf{XGLM} & \textbf{BLOOM} & \textbf{Mistral} & \textbf{Llama 2} & \textbf{Latxa} & \textbf{mGPT} & \textbf{Llama 2} & \textbf{Latxa} & \textbf{Mixtral} & \textbf{Yi} & \textbf{Llama 2} & \textbf{Latxa} \\
 & \textbf{Turbo} & \textbf{Turbo} & \textbf{7B} & \textbf{7B} & \textbf{7B} & \textbf{7B} & \textbf{7B} & \textbf{13B} & \textbf{13B} & \textbf{13B} & \textbf{8x7B} & \textbf{34B} & \textbf{70B} & \textbf{70B} \\ \midrule
Admin staff 2022 & 46.98 & \underline{\textbf{72.70}} & 22.70 & 22.99 & \textbf{34.05} & 26.58 & 31.18 & 22.84 & 30.89 & \textbf{43.53} & 41.24 & 41.24 & 37.93 & \textbf{56.03} \\
Support staff 2022 & 49.40 & \underline{\textbf{82.13}} & 25.90 & 24.50 & 32.13 & 31.93 & \textbf{43.57} & 26.91 & 32.93 & \textbf{52.61} & 47.79 & 41.37 & 34.14 & \textbf{59.84} \\
Admin assistant 2022 & 47.35 & \underline{\textbf{77.62}} & 24.39 & 26.83 & 36.30 & 31.28 & \textbf{38.16} & 26.26 & 33.14 & \textbf{49.35} & 45.34 & 41.89 & 39.89 & \textbf{57.53} \\
General questions 2022 & 45.41 & \underline{\textbf{74.09}} & 22.27 & 21.98 & \textbf{36.39} & 31.59 & 35.95 & 24.02 & 31.00 & \textbf{49.64} & 45.12 & 40.03 & 37.70 & \textbf{59.24} \\
\textbf{Public Office} & 47.29 & \underline{\textbf{76.64}} & 23.82 & 24.08 & 34.72 & 30.35 & \textbf{37.22} & 25.01 & 31.99 & \textbf{48.78} & 44.87 & 41.13 & 37.42 & \textbf{58.16} \\ \midrule
Bilbao Council 2022 & 42.38 & \underline{\textbf{71.75}} & 22.70 & 26.67 & 33.17 & 30.63 & \textbf{36.03} & 24.60 & 28.89 & \textbf{43.81} & 40.00 & 36.83 & 32.38 & \textbf{50.63} \\
Gasteiz Council 2021 & 40.83 & \underline{\textbf{67.50}} & 26.94 & 24.44 & 22.50 & 33.33 & \textbf{34.44} & 25.00 & 24.72 & \textbf{47.22} & 36.94 & 36.39 & 29.17 & \textbf{49.44} \\
\textbf{City Council} & 41.61 & \underline{\textbf{69.63}} & 24.82 & 25.56 & 27.84 & 31.98 & \textbf{35.24} & 24.80 & 26.81 & \textbf{45.52} & 38.47 & 36.61 & 30.78 & \textbf{50.04} \\
\midrule
Admin staff 2019 & 48.30 & \underline{\textbf{79.56}} & 26.25 & 21.64 & 33.47 & 29.66 & \textbf{37.68} & 23.85 & 29.66 & \textbf{54.91} & 44.89 & 43.09 & 34.27 & \textbf{60.12} \\
Admin assistant 2019 & 47.33 & \underline{\textbf{81.11}} & 24.00 & 24.67 & 36.89 & 28.22 & \textbf{42.44} & 27.56 & 33.33 & \textbf{58.89} & 45.33 & 48.44 & 40.00 & \textbf{64.44} \\
Library assistant 2019 & 45.41 & \underline{\textbf{79.63}} & 22.87 & 26.04 & \textbf{37.73} & 25.04 & 32.22 & 25.38 & 29.72 & \textbf{48.58} & 42.24 & 43.91 & 35.06 & \textbf{56.26} \\
Law 2019 & 36.14 & \underline{\textbf{69.29}} & 21.71 & 24.00 & 31.71 & 28.71 & \textbf{33.14} & 22.29 & 30.14 & \textbf{42.14} & 38.86 & 36.57 & 30.00 & \textbf{48.14} \\
Economics 2019 & 37.61 & \underline{\textbf{73.22}} & 23.08 & 23.36 & 31.34 & 25.93 & \textbf{31.91} & 25.36 & 30.48 & \textbf{44.73} & 41.31 & 36.75 & 34.47 & \textbf{52.99} \\
Business Admin 2019 & 43.21 & \underline{\textbf{73.21}} & 19.64 & 23.21 & 29.29 & 27.86 & \textbf{40.71} & 23.93 & 30.36 & \textbf{53.21} & 42.50 & 39.64 & 32.86 & \textbf{56.07} \\
Auxiliary staff 2019 & 45.25 & \underline{\textbf{80.75}} & 26.50 & 24.25 & 35.25 & 27.50 & \textbf{41.75} & 24.75 & 32.25 & \textbf{54.00} & 46.00 & 47.00 & 39.00 & \textbf{59.50} \\
Admin Tech. School (A) 2019 & 39.91 & \underline{\textbf{73.39}} & 24.03 & 24.46 & 34.19 & 27.18 & \textbf{34.05} & 22.75 & 29.33 & \textbf{46.78} & 41.20 & 39.34 & 34.33 & \textbf{53.08} \\
Admin Tech. School (B) 2019 & 47.75 & \underline{\textbf{75.79}} & 22.37 & 24.54 & \textbf{36.23} & 29.05 & 34.89 & 24.37 & 29.38 & \textbf{49.92} & 42.24 & 41.07 & 32.72 & \textbf{55.76} \\
\textbf{University} & 43.43 & \underline{\textbf{76.22}} & 23.38 & 24.02 & 33.38 & 27.68 & \textbf{37.29} & 24.47 & 30.52 & \textbf{50.35} & 42.73 & 41.76 & 34.75 & \textbf{56.26} \\
\midrule
Admin staff 2023 & 42.37 & \underline{\textbf{61.44}} & \textbf{28.81} & 22.46 & 25.00 & 23.31 & 26.27 & 27.54 & 22.46 & \textbf{36.86} & 36.86 & 34.75 & 22.46 & \textbf{43.64} \\
Health assistant 2023 & 34.73 & \underline{\textbf{57.49}} & 22.16 & 27.54 & 17.37 & \textbf{29.94} & 21.56 & 29.94 & 22.16 & \textbf{35.93} & 31.74 & 33.53 & 20.36 & \textbf{36.53} \\
Admin assistant 2023 & 37.58 & \underline{\textbf{60.00}} & 25.45 & 22.42 & 26.67 & 27.27 & \textbf{28.48} & 29.09 & 23.03 & \textbf{30.30} & 31.52 & \textbf{35.15} & 23.03 & 32.73 \\
Hospital porter 2023 & 33.13 & \underline{\textbf{63.19}} & 22.70 & 25.77 & 26.38 & 28.22 & \textbf{32.52} & 33.13 & 25.15 & \textbf{33.74} & 35.58 & 30.06 & 26.99 & \textbf{39.26} \\
Medical staff 2023 & 39.14 & \underline{\textbf{60.59}} & 26.01 & 21.45 & 23.59 & 24.66 & \textbf{25.74} & 28.15 & 22.25 & \textbf{36.19} & 33.24 & 35.12 & 27.61 & \textbf{39.68} \\
Service operator 2023 & 36.64 & \underline{\textbf{58.02}} & 25.19 & 19.85 & \textbf{29.01} & 28.24 & 28.24 & 28.24 & 22.90 & \textbf{32.06} & 34.35 & 32.06 & 20.61 & \textbf{42.75} \\
Superior technician 2023 & 40.19 & \underline{\textbf{58.88}} & \textbf{30.84} & 24.92 & 25.86 & 28.35 & 26.17 & 28.04 & 26.48 & \textbf{37.38} & 33.33 & 33.33 & 27.41 & \textbf{36.76} \\
Misc (cook, janitor, etc.) 2023 & 38.72 & \underline{\textbf{61.65}} & 28.57 & 24.06 & 24.44 & 26.69 & \textbf{29.70} & 28.95 & 24.06 & \textbf{38.72} & 33.83 & 33.46 & 22.18 & \textbf{39.47} \\
Admin assistant 2008 & 37.25 & \underline{\textbf{57.96}} & 24.57 & 22.41 & 29.98 & 25.81 & \textbf{28.13} & 23.49 & 27.82 & \textbf{31.68} & 28.59 & 33.08 & 29.52 & \textbf{44.05} \\
Admin staff 2008 & 36.63 & \underline{\textbf{59.22}} & 24.87 & 24.06 & \textbf{32.49} & 28.48 & 30.35 & 24.47 & 28.61 & \textbf{37.43} & 35.29 & 35.56 & 31.95 & \textbf{44.12} \\
Hospital porter 2008 & 42.08 & \underline{\textbf{70.31}} & 26.96 & 28.05 & 34.24 & \textbf{34.43} & 32.24 & 26.23 & 29.33 & \textbf{42.26} & 40.98 & 41.71 & 35.34 & \textbf{54.28} \\
Auxiliary nurse 2008 & 52.77 & \underline{\textbf{81.69}} & 25.54 & 28.92 & \textbf{35.23} & 30.46 & 32.46 & 29.08 & 34.31 & \textbf{44.46} & 43.69 & 43.38 & 37.38 & \textbf{56.00} \\
Nurse 2008 & 51.90 & \underline{\textbf{81.00}} & 25.60 & 31.40 & 36.30 & 32.10 & \textbf{36.50} & 32.40 & 33.20 & \textbf{43.60} & 41.50 & 46.00 & 34.40 & \textbf{57.30} \\
Family doctor 2008 & 43.83 & \underline{\textbf{73.09}} & 25.01 & 25.52 & 30.65 & 27.69 & \textbf{31.25} & 24.50 & 29.54 & \textbf{37.45} & 37.08 & 39.34 & 31.85 & \textbf{49.24} \\
\textbf{Health System} & 40.50 & \underline{\textbf{64.61}} & 25.88 & 24.92 & 28.37 & 28.26 & \textbf{29.39} & 28.09 & 26.52 & \textbf{36.66} & 35.54 & 36.18 & 27.94 & \textbf{43.78} \\
\midrule
\textbf{Average} & 42.42 & \underline{\textbf{70.22}} & 24.75 & 24.57 & 30.76 & 28.63 & \textbf{33.32} & 26.31 & 28.54 & \textbf{43.19} & 39.26 & 38.62 & 31.55 & \textbf{50.07} \\
\midrule

\end{tabular}
\end{adjustbox}
\caption{Detailed results on EusExams tests and categories (in \textbf{bold}). Best results in each compute class are in \textbf{bold}. Best overall results are \underline{underlined}.}
\label{tab:results-eusexams-detailed}
\end{table*}

\end{document}